\documentclass[twoside,11pt]{article}

\usepackage{jmlr2e}

\usepackage[utf8]{inputenc} 
\usepackage[T1]{fontenc}    
\usepackage{hyperref}       
\usepackage{url}            
\usepackage{booktabs}       
\usepackage{amsfonts}       
\usepackage{nicefrac}       
\usepackage{microtype}      
\usepackage{times}
\usepackage{graphicx} 
\usepackage{subcaption}
\usepackage{tabularx}
\usepackage{bbm}
\usepackage{url}
\usepackage{amsmath}
\usepackage{bbm}
\usepackage{caption}
\usepackage{float}
\usepackage{amssymb}
\usepackage{grffile}
\usepackage{epstopdf}
\usepackage[colorinlistoftodos]{todonotes}
\usepackage{multirow}

\usepackage{color}

\usepackage[normalem]{ulem}

\newcommand{\hide}[1]{}

\def\b{\ensuremath\boldsymbol}

\usepackage[numbers]{natbib}

\begin{document}

\title{Affective Manifolds: Modeling Machine's Mind to Like, Dislike, Enjoy, Suffer, Worry, Fear, and Feel Like A Human}

\author{\name Benyamin Ghojogh \email bghojogh@uwaterloo.ca 
}
	
\editor{ }

\maketitle

\begin{abstract}
After the development of different machine learning and manifold learning algorithms, it may be a good time to put them together to make a powerful mind for machine. In this work, we propose affective manifolds as components of a machine's mind. Every affective manifold models a characteristic group of mind and contains multiple states. We define the machine's mind as a set of affective manifolds. We use a learning model for mapping the input signals to the embedding space of affective manifold. Using this mapping, a machine or a robot takes an input signal and can react emotionally to it. We use deep metric learning, with Siamese network, and propose a loss function for affective manifold learning. We define margins between states based on the psychological and philosophical studies. Using triplets of instances, we train the network to minimize the variance of every state and have the desired distances between states. We show that affective manifolds can have various applications for machine-machine and human-machine interactions. Some simulations are also provided for verification of the proposed method. It is possible to have as many affective manifolds as required in machine's mind. More affective manifolds in the machine's mind can make it more realistic and effective. This paper opens the door; we invite the researchers from various fields of science to propose more affective manifolds to be inserted in machine's mind.
\end{abstract}
\hfill\break


\medskip

\section{Introduction}

Now that many machine learning algorithms have been proposed, it is time to combine them and put them together to make a powerful machine's mind. On the one hand, this especially makes sense when we notice that human is a complex machine so it may be worth attempting to model the mind of this machine. 
On the other hand, the development of technology and science encourages us to improve machines. We never know; the machines may coexist with humans, as intelligent semi-human robots, in the near future. 

Modeling machine's mind is related to affective computing. 
Affective computing \citep{picard2000affective} is generally referred to two broad research areas:
(1) either emotion recognition from signals such as facial images, or (2) how a machine feels and has emotions. The concept of machine's mind and emotions, addressed in this paper, lies in the latter research area of affective computing. Moreover, machine's feeling can be used in human-computer and computer-computer interactions; this will be discussed further in Section \ref{section_applications}.

In this paper, we propose and define the concept of \textit{affective manifolds} for modeling the mind of machine/robot. The proposed tool can be used in a machine to possess a mind and feel like a human. The machine, or a robot, needs to have a mind and feel like a human so it seems realistic to humans. Affective manifolds can provide this ability to machines. 
A machine's mind can be considered as a set of affective manifolds (this will be defined formally in Section \ref{section_affective_manifolds}).
A machine's mind can have as many affective manifolds as required.
The more affective manifolds are designed gradually by researchers, the better and more complete the machine's mind can become. Eventually, the machine's mind can become complex enough like a human's mind. Consequently, we invite the researchers to design more and better affective manifolds to insert in the machine's mind.

The remainder of this paper is organized as follows. 
We define and introduce the machine's mind and affective manifolds, as well as some examples, in Section \ref{section_affective_manifolds}. One possible machine learning algorithm for learning the affective manifolds is proposed in Section \ref{section_affective_manifold_learning}. Section \ref{section_applications} introduces some possible applications of the proposed affective manifolds. Simulations are provided in Section \ref{section_simulations}. Finally, Section \ref{section_conclusion} concludes the paper and enumerates the possible future directions. 

\section{Machine's Mind and Affective Manifolds}\label{section_affective_manifolds}

\subsection{Modeling Alive Machine}

Imitating any alive creature, we can model a machine. For example, we can have vegetable-like, animal-like, and human-like machines which can play the roles of robots as vegetables (plants), animals, and humans. These machines have some advantages and benefits to biological organisms. For example, an alive vegetable-like machine does not wilt. An alive animal-like machine, such as a dog robot, does not dye or can be taught very fast. 

An alive human-like machine is more robust to damages and can work in dangerous environments. Moreover, they do not get tired from working. Although, as we will show in this paper, we can add the ability of getting tired or having pain to the machines. It is possible to have multiple types of human-like machines with different abilities where the worker robots do not posses the ability of pain and tiredness but the living robots have these characteristics to live among humans. Nonetheless, this may raise the ethical dilemmas of racism/distinction among robots and racism/distinction between human and robot, which can be serious problems in the future. This reminds us that affective computing usually faces ethical challenges \citep{picard2003affective}.  

\subsection{Affective Manifolds}

Diffusion MRI and fMRI experiments \cite{jones2010diffusion} have shown that different parts of brain are activated based on various mood characteristics, such as joy, suffer, love, fear, etc.
Accordingly, we can define mind as a set of affective manifolds, each of which corresponds to a group of characteristics of mind. 
This models the different parts of machine's mind.
The affective manifolds play the role of latent variables for characteristics of mind. 
Some example effective manifolds will be mentioned in Section \ref{examples}.

\begin{definition}[Affective Manifold]
The characteristics of mind can be grouped into multiple categories.
The affective manifold $\mathcal{M}$ is a manifold corresponded to a characteristic group. The machine learns this manifold in order to distinguish between the states of that characteristic group. If $\b{x}_1$ and $\b{x}_2$ correspond to two different states $\mathcal{S}_1$ and $\mathcal{S}_2$ in a characteristic group, represented by the affective manifold $\mathcal{M}$, they are expected to fall away from each other on the manifold, compared to their own states:
\begin{equation}
\begin{aligned}
(\b{x}_1 \in \mathcal{S}_1 \in \mathcal{M}) \wedge (\b{x}_2, \b{x}_3 \in \mathcal{S}_2 &\in \mathcal{M}) \wedge (\mathcal{S}_2 \cap \mathcal{S}_2 = \varnothing) \\
& \implies \mathbb{E}\big[\|\b{x}_2 - \b{x}_3\|_2\big] \ll \mathbb{E}\big[\|\b{x}_2 - \b{x}_1\|_2\big],
\end{aligned}
\end{equation}
where $\mathbb{E}[.]$ denotes the expected value. 
The affective manifold, corresponded to a characteristic group, is a collection of $s$ disjointed states:
\begin{align}
&\mathcal{M} := \bigcup_{i=1}^s \mathcal{S}_i, \\
&\mathcal{S}_i \cap \mathcal{S}_j = \varnothing, \quad \forall i, j \in \{1, \dots, s\},
\end{align}
where it is theoretically possible to have $s = \infty$.
In other words, an affective manifold is partitioned into its states. 
\end{definition}
An example affective manifold is the love manifold containing the states of ``love'', ``like'', ``dislike'', and ``hate''. 
More detailed examples for affective manifold will be provided in Section \ref{examples}.

\begin{definition}[Affective Subspace]
If we use a linear method for learning an affective manifold, the affective manifold is linear. In this case, the affective manifold is reduced to an affective subspace. 
The affective subspace, which is a special case of the affective manifold, can be modeled as a linear column-space of a projection matrix $\b{U} \in \mathbb{R}^{d \times p}$ from a $d$-dimensional input space to a $p$-dimensional subspace, where $p \leq d$. Therefore, the affective subspace belongs to the Grassmannian manifold $\mathcal{G}(p,d)$, which is a space of all $p$-dimensional linear subspaces of the $d$-dimensional vector space. 
The Grassmannian manifold $\mathcal{G}(p,d)$ can be seen as the quotient space of the Stiefel manifold $\mathcal{S}t(p,d)$ \cite{absil2009optimization}:
\begin{align}
\mathcal{G}(p,d) := \mathcal{S}t(p,d) / \mathcal{S}t(p,p),
\end{align}
where the Stiefel manifold is defined as the set of orthogonal matrices as:
\begin{align}
\mathcal{S}t(p,d) := \{\b{U} \in \mathbb{R}^{d \times p}\, |\, \b{U}^\top \b{U} = \b{I}\}.
\end{align}
\end{definition}
Any linear dimensionality reduction method, such as Fisher discriminant analysis \cite{ghojogh2019fisher}, can be used to learn an affective subspace. 

\subsection{Machine's Mind}

\begin{figure*}[!t]
\centering
\includegraphics[width=\textwidth]{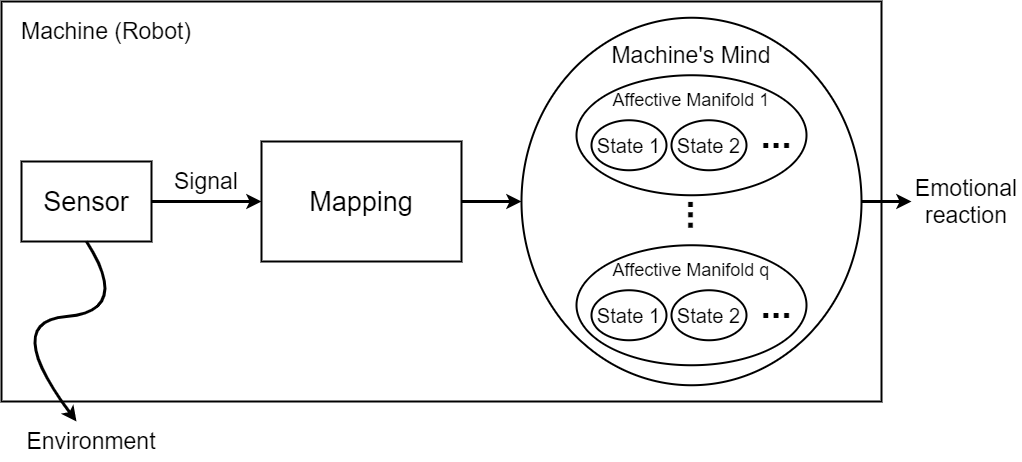}
\caption{The pipeline of sensing the signal by the machine/robot and reacting emotionally to it.}
\label{figure_pipeline}
\end{figure*}

\begin{definition}[Machine's Mind]
The characteristics of mind, in terms of mood, can be grouped into multiple categories.
Let the manifold $\mathcal{M}_i$ denote the $i$-th manifold corresponded to the $i$-th characteristic group. The machine's mind, denoted by $\mathcal{Q}$, is a set of $q$ affective manifolds:
\begin{align}\label{equation_machine_mind}
\mathcal{Q} := \bigcup_{i=1}^q \mathcal{M}_i,
\end{align}
where every affective manifold is responsible for a characteristic group, and it is theoretically possible to have $q = \infty$.
\end{definition}
An illustration of parts of machine's mind is in Fig. \ref{figure_pipeline}.
Eq. (\ref{equation_machine_mind}) simulates the brain of human whose different parts are activated for different tasks, according to the diffusion MRI and fMRI experiments \cite{jones2010diffusion}.

\begin{remark}
In a machine's mind, it is possible to have overlapping affective manifolds where some states are shared between some characteristic groups. Therefore, for two states from two affective manifolds, i.e., $\mathcal{S}_1 \in \mathcal{M}_1$ and $\mathcal{S}_2 \in \mathcal{M}_2$, we can have either $\mathcal{S}_1 \cap \mathcal{S}_2 = \varnothing$ or $\mathcal{S}_1 \cap \mathcal{S}_2 \neq \varnothing$. This means we can have either $\mathcal{M}_1 \cap \mathcal{M}_2 = \varnothing$ or $\mathcal{M}_1 \cap \mathcal{M}_2 \neq \varnothing$ for every two affective manifolds $\mathcal{M}_1$ and $\mathcal{M}_2$ in a machine's mind.
\end{remark}
An example for two non-overlapping affective manifolds is where one manifold contains the states ``frown'', ``neutral lips'', ``smile'', and ``laugh'', and the other manifold includes the states ``feared'', ``worried'', ``enjoy'', and ``laugh''. The state ``laugh'' is shared between the two characteristic groups of these affective manifolds.

\subsection{Example Affective Manifolds}\label{examples}

We can define various affective manifolds. As many effective manifolds as required can be defined and used. Some example affective manifolds are provided in the following. 

\subsubsection{Affective Love Manifold}

An example affective manifold is the affective love manifold to model the loving characteristic group. This manifold can include the states ``hate'', ``dislike'', ``like'', and ``love''. This enables the machine to love, like, dislike, and hate different situations, objects, humans, or other machines.

\subsubsection{Affective Emotion Manifold}

Another example affective manifold is the affective emotion manifold to model the emotion characteristic group. This manifold can include the states ``angry'', ``sad'', ``neutral'', ``glad'', and ``excited''. This enables the machine to have emotions and feelings like a human. 

\subsubsection{Affective Joy Manifold}

Another example for affective manifolds is the affective joy manifold containing the states ``suffered'', ``feared'', ``worried'', ``enjoying'', ``relaxed'', and ``bored''. 
This enables the machine to have different states/levels of joy in different circumstances. 

For two reasons, it may be plausible to insert the ability of feeling pain or suffering into the mind of machine. In the first glance, pain may be a negative characteristic which is not desired to have in an ideal machine. However, biology has shown us that pain has saved humans and living creatures to survive by understanding and preventing danger, sickness, and injuries. So, pain can prevent robots from being destroyed.
Another benefit of pain/suffer is that it gives meaning to happiness; otherwise, if all feelings were happiness, there was no any concept for happiness. In other words, pain and suffer, as well as happiness and joy, give living creatures a mixture of feelings for life. For these reasons, pain may be plausible.

It is also noteworthy that the adversarial attack \cite{chakraborty2018adversarial} and/or computer viruses \cite{kephart1993measuring} can model disease for machines. Anti-viruses can be developed for defeating machine's diseases. Developing defense algorithms for robustness to adversarial attack can also defeat machine's diseases.

\subsubsection{Affective Belief Manifold}

Another example for affective manifolds is the affective belief manifold where the machine believes in one or several of the existing belief systems. Like humans, the machine may change its mind in its belief gradually over time. The belief manifold may have $b$ existing belief systems, an agnosticism state, and an atheism state. It is also possible to have other belief systems and having agnosticism and atheism associated to some sets of beliefs.
The number of belief systems in the machine's mind may be infinite. 
The affective belief system can enable machines to have various beliefs like humans. 

\section{Affective Manifold Learning}\label{section_affective_manifold_learning}

We can use any dimensionality reduction and manifold learning model, such as FDA \cite{ghojogh2019fisher}, locally linear embedding \cite{roweis2000nonlinear,ghojogh2020locally}, and deep metric learning \cite{ghojogh2022spectral}, for learning every affective manifold. 
It is also possible to use the mathematical techniques such as differential geometry \cite{kuhnel2015differential} or Riemannian optimization \cite{absil2009optimization} to learn the affective manifolds. In this section, we provide one of the possible algorithms for affective manifold learning. 

\subsection{Dimensionality of the Affective Manifold}

We can use different learning methods for different manifolds depending on their complexities.
The dimensionality of every affective manifold or subspace, denoted by $p$, depends on the complexity of that characteristic group. The more complex the characteristic is, the larger dimensionality of manifold is required. Both complexities of states and the number of states have impact on complexity: 
\begin{itemize}
\item The more number of states a characteristic group has, the more complex the affective manifold is, so a larger dimensionality is required for the affective manifold.
\item The more complex the states of a characteristic group are, the more complex the affective manifold is, so a larger dimensionality is required for the affective manifold.
\end{itemize}

\subsection{The Learning Model}

The goal is to learn a nonlinear mapping $f$, with learnable parameters $\theta$, from a $d$-dimensional input space to a $p$-dimensional embedding space:
\begin{align}\label{equation_mapping}
f(\theta): \mathbb{R}^d \rightarrow \mathbb{R}^p.
\end{align}
The $d$-dimensional data can be any signal captures by any sensor of the machine/robot. The input signal stimulates one of the states in the affective manifold, the same way as an input signal to human's brain stimulates a state in human's mind (for example, we can get sad by looking at a picture or listening to a piece of music). 
The input signals are measured by any sensor of the machine, where the sensors can model the five senses of a human. For example, the input signals can be images, observed by the cameras (eyes) of robot, sound signals heard by microphones, or any other signals. 
The $p$-dimensional embedding space is the affective manifold. Therefore, the mapping (\ref{equation_mapping}) is a mapping from the input signal to the affective manifold in the machine's mind. 
The pipeline of mapping the input signals to the affective manifolds in the machine's mind is illustrated in Fig. \ref{figure_pipeline}.

We can learn an affective manifold using deep metric learning \cite{ghojogh2022spectral}. One useful network for deep metric learning is the Siamese network \cite{bromley1993signature} containing two or three networks sharing their weights. The mapping $f$ can be a Siamese neural network whose weights are the learnable parameters $\theta$.

\subsection{Margins Between the States}

Based on the psychological experiments, philosophical studies, and biological facts, we can define the desired relative distances (or margins) between the states of an affective manifold. For this, in the embedding space of the manifold, we define the relative margins between the states of that manifold. Note that the actual distances are not important but the relative distances matter. Scaling up or down the actual distances can only have impact on the speed of convergence of training. In addition to the scale of distances, the rotation and mirroring are not important in manifold learning because the relative distances are preserved under these transformations. 

Figure \ref{figure_love_manifold} depicts an example for imagination of the desired relative distances between the states of an affective manifold. Section \ref{section_simulations} will provide examples for the relative distances of states in simulations.
Note that defining the margins between the states of an affective manifold requires empirical social experiments, biological facts, psychological studies, or philosophical arguments. Some psychological and philosophical analyses used for determining the margins will be discussed in Section \ref{section_simulations}.

If an affective manifold has $s$ states, we define a symmetric margin matrix $\b{M} \in \mathbb{R}_+^{s \times s}$ whose elements are the desired margins between the states. Let $m_{i,j}$ denote the $(i,j)$-th element of the matrix $\b{M}$, i.e., the desired margin between the $i$-th and the $j$-th states of the affective manifold. 

\subsection{Triplets}

For learning an affective manifold, we can learn an embedding space where the variance of instances within every state is minimized and the inter-state distances are fixed to the desired distances. For this goal, we prepare mini-batches of $d$-dimensional anchor-positive-negative triplets. In every triplet, the anchor and positive instances belong to the same state but the negative instance is from another state. 

In every mini-batch, let the $j$-th $d$-dimensional anchor, positive, and negative instances be denoted by $\b{x}^j_a \in \mathbb{R}^d$, $\b{x}^j_p \in \mathbb{R}^d$, and $\b{x}^j_n \in \mathbb{R}^d$, respectively. If the bach size is denoted by $b$, every mini-batch is $\{(\b{x}^j_a, \b{x}^j_p, \b{x}^j_n)\}_{j=1}^b$. The embeddings of triplets in the affective manifold are $f(\b{x}^j_a) \in \mathbb{R}^p$, $f(\b{x}^j_p) \in \mathbb{R}^p$, and $f(\b{x}^j_n) \in \mathbb{R}^p$.
We denote the state of the instance $\b{x}$ by $s(\b{x})$; hence, the states of a triplet are $s(\b{x}^j_a)$, $s(\b{x}^j_p)$, and $s(\b{x}^j_n)$.

\subsection{Training the Affective Manifold}

Inspired by the triplet loss \cite{schroff2015facenet} and the contrastive loss \cite{hadsell2006dimensionality}, we propose a loss function for learning an affective manifold.
We define the following loss function for training the affective manifold:
\begin{align}\label{equation_total_loss}
\theta := \arg \min_{\theta} \mathcal{L} = \arg \min_{\theta} \big( \lambda_p \mathcal{L}_p + \lambda_n \mathcal{L}_n \big),
\end{align}
where $\lambda_p > 0$ and $\lambda_n > 0$ are the weighting parameters controlling the relative importance of the positive loss $\mathcal{L}_p$ and the negative loss $\mathcal{L}_n$. The functions $\mathcal{L}_p$ and $\mathcal{L}_n$ are the loss functions for the positive and negative pairs:
\begin{align}
&\mathcal{L}_p := \frac{1}{b} \sum_{j=1}^b \big\| f(\b{x}^j_a) - f(\b{x}^j_p) \big\|_2^2, \label{equation_positive_loss} \\
&\mathcal{L}_n := \frac{1}{b} \sum_{j=1}^b \Big( \big\| f(\b{x}^j_a) - f(\b{x}^j_n) \big\|_2 - m_{s(\b{x}^j_a), s(\b{x}^j_n)} \Big)^2, \label{equation_negative_loss}
\end{align}
where $\|.\|_2$ denotes the $\ell_2$ norm.

The loss function (\ref{equation_positive_loss}) is the mean squared error of distances between the anchor and positive of triplets in a mini-batch. 
The loss function (\ref{equation_negative_loss}) is the mean squared error of difference of the desired margin and the current distance between the anchor and negative of triplets in a mini-batch. 
Therefore, on the one hand, the overall loss function (\ref{equation_total_loss}) reduces the distances between instances of every state so that the veriance of every state is reduced. On the other hand, this loss function makes the distances between every two states equal to the desired margin between those two states.

It is noteworthy that machine's every affective manifold can change gradually over time using transfer learning \cite{weiss2016survey}. For this, we further train the already trained network using new triplets from new input signals. This transfer learning simulates experiencing new things by the machine. This is inspired by the gradual change of human's mind and characteristics over time by experiencing.

\subsection{Inferring the State of an Input Signal}

For every $d$-dimensional input signal, the machine shall have one of the states of an affective manifold. 
After training the network, we pass the input $\b{x} \in \mathbb{R}^d$ to the network and we get its embedding $f(\b{x}) \in \mathbb{R}^p$ at the output of network. We can have various approaches for inferring the state of the input signal in the embedding space. One possible approach is explained in the following. 
Let the mean of embeddings of the training instances in the $i$-th state be denoted by $\b{s}_i \in \mathbb{R}^p$. 
The state of the input signal can be determined as:
\begin{align}\label{equation_infer_state}
s(\b{x}) := \arg \min_{i \in \{1, \dots, s\}} \|f(\b{x}) - \b{s}_i\|_2.
\end{align}
Using other distance metrics such as the Mahalanobis distance or other norms is also possible. 
By inferring the state, the machine's mind becomes able to possess a state of that characteristic group in reaction to the input signal. For example, the machine becomes happy or sad by looking at a specific picture or listening to a specific sentence/music. 

\section{Some Applications of Affective Manifolds}\label{section_applications}

Affective manifolds can have various applications. Some example applications are provided in the following. 

\subsection{Machine-Machine Interaction}

If two machines have minds, containing affective manifolds, they can interact with each other and react emotionally accordingly. For example, they can talk to one another and see each other and react to the chat or the body language of the other machine (if the machines are implemented in robots). In this meaning, two machines (or robots) can interact with each other similar to how two humans interact, if they have sufficient number of affective manifolds in their minds. For instance, two machines can fall in love with each other, dislike one another, or get angry at each other (e.g., if one of them verbally insults the other one and the other machine gets sad). 

\subsection{Human-Machine Interaction}

If a machine or robot has a mind with affective manifolds, it can interact with human(s) or living creatures. This also opens the gate for forming positive and negative emotions between human and machine. For example, the robot and human can interact and get happy, get sad, or even fall in live with each other. Sci-fi movies have tried to show the possibility of such situations. For example, in the movie ``Her'' directed by Spike Jonze, a man falls in love with the operating system of his computer. Another example is the movie ``Ex Machina'', directed by Alex Garland, in which a human falls in love with a robot. 

Any positive and negative emotions and feedback can happen between human and machine. It is possible to have these emotions under control if the number of states in the affective manifolds are controlled (for example, if we remove some controversial states from the machine's mind). Forming emotions between human and machine may open the gate for them to even fall in love with each other \cite{samani2010towards}; some researchers have even gone a step further and raised the possibility of marriage of human and robot in the future \cite{levy2007love}. Obviously, this raises serious ethical challenges which need to be addressed \cite{sullins2012robots}; however, we should note that affective computing is always faced with ethical difficulties \citep{picard2003affective}. 

\subsection{Human-Human Interaction}

The affective manifolds can also be useful for human-human interaction. This might seem strange as there is not any machine in the sides of this interaction. However, note that it is possible to model a human's mind as a machine's mind. This simplifies (reduces) the complex human's mind to a simpler mind but with sufficiently good imitation. The modeled mind can almost produce same behavior as the human's mind if it has enough affective manifolds with enough number of states and is trained well enough. 

An example usage of this application is using this model in matchmakers for dating apps. For example, the behavior of the human users, based on their characteristics and their likes/dislikes, can be used to train an affective manifold with states ``match'' (``like'') and ``non-match'' (``dislike'') for every user. Then, for each user, the characteristics of other users are fed as input signals to the learning model to see where every other user falls in the affective manifold. The user with the closest position in the state of ``match'' (``like'') will be the best match for that user (see Eq. (\ref{equation_infer_state})).

\section{Simulations}\label{section_simulations}

In this section, we provide some simulation examples of affective manifold learning. The codes of simulation for this paper can be found in the following link:
\url{https://github.com/bghojogh/Affective-Manifold}.

\subsection{Setting of Simulations}

We used a Siamese network for affective manifold learning. The backbone network included two convolutional layers followed by two fully connected layers. The Parametric Rectified Linear Unit (PReLU) activation functions, max pooling, and dropout were used in the network. The batch size was set to $b=32$ and the embedding dimensionality of affective manifold was set to $p=2$ for better visualization of results. We set $\lambda_p = \lambda_n = 1$ to have equal weights in the loss function (\ref{equation_total_loss}). For every experiment, we trained the network for ten epochs. For the sake of simulation, we used the MNIST dataset \cite{lecun1998gradient} as the input signal where every digit is associated to one of the states. 

\subsection{Affective Love Manifold}

The affective love manifold can have the states ``hate'', ``dislike'', ``like'', and ``love''. Based on two different physiological observations, we can define the margins either linearly or nonlinearly. 

\subsubsection{Linear Affective Love Manifold}

\begin{figure*}[!t]
\centering
\includegraphics[width=0.8\textwidth]{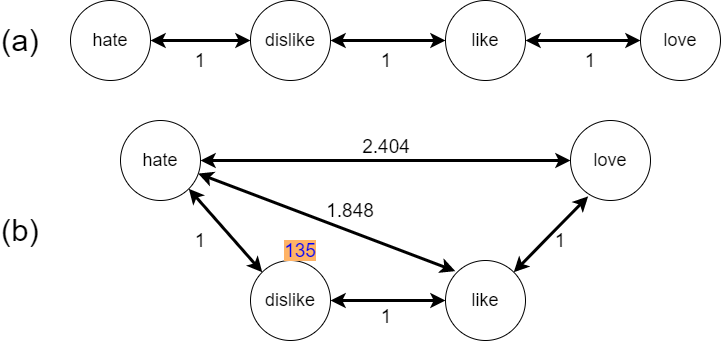}
\caption{The desired margins in the (a) linear and (b) nonlinear versions of love manifold. The numbers on lines are the desired distances and the number with a colored background is the desired angle in degrees.}
\label{figure_love_manifold}
\end{figure*}

In the first simple glance, we can say that ``hate'' is worse than ``dislike'', ``dislike'' is worse than ``like'', and ``like'' is worse than ``love''. Therefore, we can possibly see their relations over a line in a linear relationship, as illustrated in Fig. \ref{figure_love_manifold}-a.
Therefore, if ``hate'', ``dislike'', ``like'', and ``love'' are the first to fourth states, the relative margins can be defined as:
\begin{align}
\b{M} := 
\begin{bmatrix}
0 & 1 & 2 & 3\\
1 & 0 & 1 & 2\\
2 & 1 & 0 & 1\\
3 & 2 & 1 & 0\\
\end{bmatrix}.
\end{align}

\subsubsection{Nonlinear Affective Love Manifold}

A deeper investigation into the love-related emotions can show us that the relations of the states ``hate'', ``dislike'', ``like'', and ``love'' is not necessarily linear. Love is not that simple. In one of her novels of Hercule Poirot, Agatha Christie correctly says \cite{christie1937death}:
\begin{quote}
``Love can be a very frightening thing. That is why most great love stories are tragedies.''

\noindent\hfill Agatha Christie, ``Death on the Nile'', 1937.
\end{quote}
Humans have observed, in their daily lives, that love is not as stable as liking. It can suddenly convert to hate, unfortunately. Therefore, we can define the margins specified in Fig. \ref{figure_love_manifold}-b, where the states of ``hate'' and ``love'' are slightly curved toward one another. In this case, we can calculate relative margins based on the desired angles and relative positions of states:
\begin{align}\label{equation_M_nonlinear_love_manifold}
\b{M} := 
\begin{bmatrix}
0 & 1 & 1.848 & 2.404 \\
1 & 0 & 1 & 1.848 \\
1.848 & 1 & 0 & 1 \\
2.404 & 1.848 & 1 & 0 \\
\end{bmatrix},
\end{align}
where ``hate'', ``dislike'', ``like'', and ``love'' are the first to fourth states.

\begin{figure*}[!t]
\centering
\includegraphics[width=\textwidth]{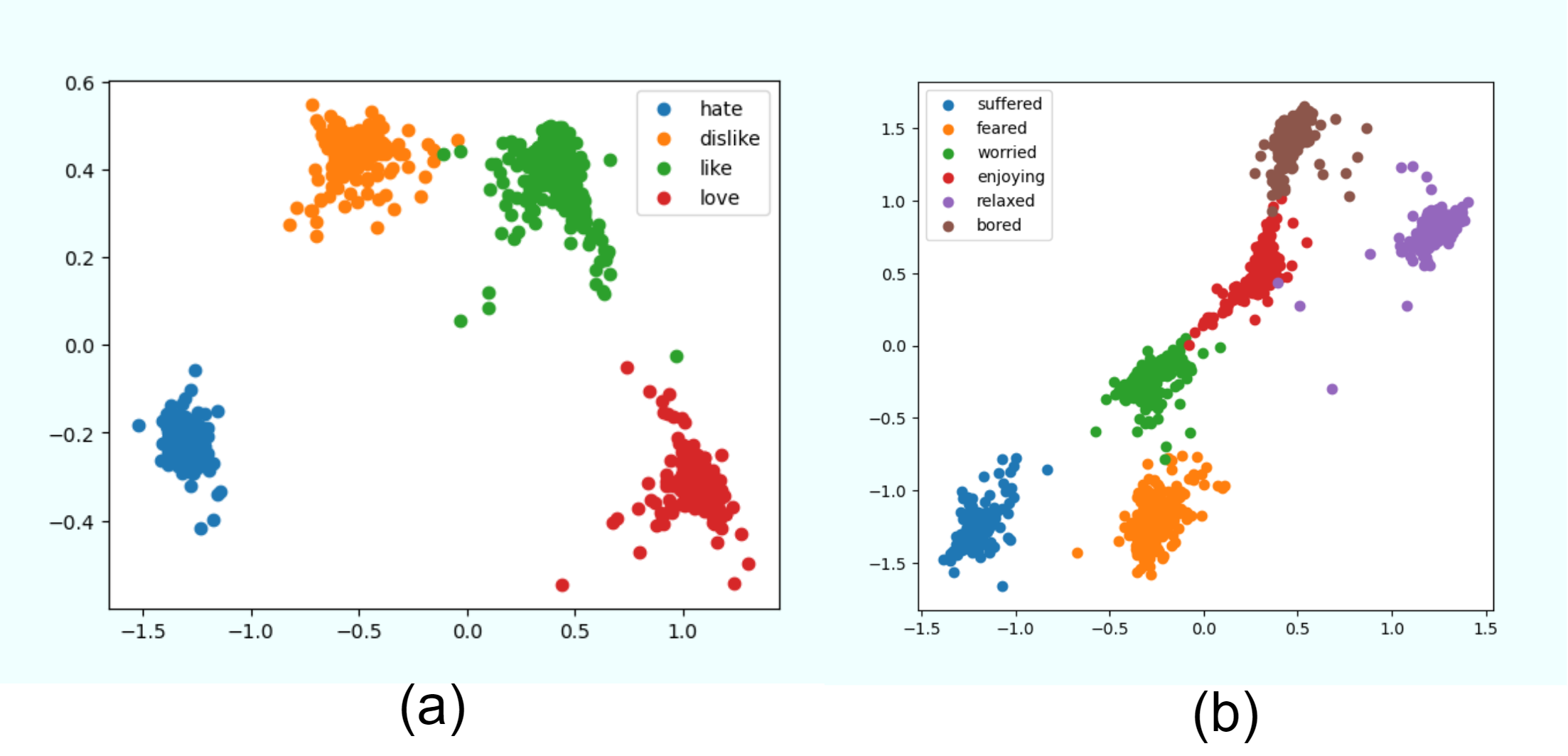}
\caption{The trained embedding spaces of (a) nonlinear affective love manifold and (b) affective joy manifold.}
\label{figure_results}
\end{figure*}

The trained embedding space for the nonlinear affective love manifold can be seen in Fig. \ref{figure_results}-a. As this figure shows, the states are correctly learned to follow the desired margins determined in Eq. (\ref{equation_M_nonlinear_love_manifold}) and shown in Fig. \ref{figure_love_manifold}-b. Note that, in manifold learning, the relative distances are important and the overall scale, rotation, and mirroring are not important.

\subsection{Affective Joy Manifold}

\begin{figure*}[!t]
\centering
\includegraphics[width=0.35\textwidth]{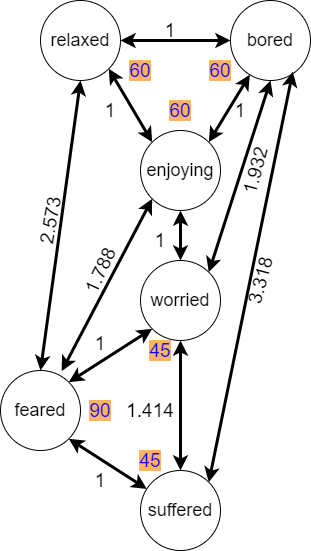}
\caption{The desired margins in the joy manifold. The numbers on lines are the desired distances and the numbers with colored background are the desired angles in degrees.}
\label{figure_joy_manifold}
\end{figure*}

The affective joy manifold models the level of joy in the machine's mind. This manifold can contain the states ``suffered'', ``feared'', ``worried'', ``enjoying'', ``relaxed'', and ``bored''. One possible way, but not the only way, to define the margins between these states is illustrated in Fig. \ref{figure_joy_manifold}. As this figure shows, suffering is the opposite of relaxing and being bored in terms of comfort. We set the state ``feared'' away from ``suffered'' and ``worried'' with the same distances. However, among ``feared'' and ``suffered'', it seems better to have ``feared'' closer to ``worried'' because suffering is more severe. Among the negative joy levels, the state ``worried'' can be the closest to the state ``enjoying'' which is positive. For the neutral joy levels, which are ``relaxed'' and ``bored'', we define the margins equal from the state ``enjoying''. 

The reason for why we put the positive state ``enjoying''  between the negative states (i.e., ``suffered'', ``feared'', and ``worried'') and the neutral states (i.e., ``relaxed'', and ``bored'') is explained in the following. Previously, humans thought that suffering and joy are the two opposite ends of the levels of joy in life. However, philosophical investigations into joy and meaning of life have shown that enjoying is merely a short period of time, between suffering and boredom, in every experience of life. Human starts from the suffer of not having something, tries and obtains it, then enjoys it for some short time, and then the human becomes bored of that. This cycle starts again for obtaining something else. In fact, boredom, and not enjoying, is the opposite of suffering in life. This was initially thought through by Gautama Buddha. In the modern philosophy, Arthur Schopenhauer has addressed this concept:
\begin{quote}
``Life swings like a pendulum backward and forward between pain and boredom.''

\noindent\hfill Arthur Schopenhauer, 1788--1860.
\end{quote}

According to the above explanations and Fig. \ref{figure_joy_manifold}, we can calculate relative margins based on the desired angles and relative positions of states:
\begin{align}\label{equation_M_joy_manifold}
\b{M} := 
\begin{bmatrix}
0 & 1 & 1.414 & 2.414 & 3.318 & 3.318 \\
1 & 0 & 1 & 1.788 & 2.573 & 2.761 \\
1.414 & 1 & 0 & 1 & 1.932 & 1.932 \\
2.414 & 1.788 & 1 & 0 & 1 & 1 \\
3.318 & 2.573 & 1.932 & 1 & 0 & 1 \\
3.318 & 2.761 & 1.932 & 1 & 1 & 0 \\
\end{bmatrix},
\end{align}
where ``suffered'', ``feared'', ``worried'', ``enjoying'', ``relaxed'', and ``bored'' are the first to sixth states.

The trained embedding space for the affective joy manifold is illustrated in Fig. \ref{figure_results}-b. As this figure shows, the states are correctly learned to follow the desired margins determined in Eq. (\ref{equation_M_joy_manifold}) and depicted in Fig. \ref{figure_joy_manifold}. Again, note that the relative distances are important and the overall scale, rotation, and mirroring do not matter. 

\section{Conclusion and Future Directions}\label{section_conclusion}

In this paper, we proposed the affective manifolds in the machine's mind. The machine's mind is a set of affective manifolds and every affective manifold is a collection of states. Each affective manifold corresponds to a characteristic group of mind. We enumerated various examples and applications of the affective manifolds. 

This work opens a door to designing machine learning and manifold learning models for learning various affective manifolds as characteristics of mind. In this work, we proposed a loss function, with the desired margins between states, for deep metric learning in affective manifold learning. Other loss functions and/or other machine learning models can be used for affective manifold learning. 

It is also possible to learn the affective manifolds using differential geometry and Riemannian optimization.
We invite researchers from various fields of science to propose more affective manifolds to include in the machine's mind. In addition, because of the concept of mind in affective manifolds, it is possible to use psychology and psychoanalysis for affective manifolds in the machine's mind.

\bibliography{ref}



\end{document}